\documentclass[letterpaper, 10 pt, conference]{ieeeconf}  

\IEEEoverridecommandlockouts                              

\usepackage{graphics} 
\usepackage{epsfig} 
\usepackage{bm} 
\usepackage{mathptmx} 
\usepackage{times} 
\usepackage{amsmath} 
\usepackage{amssymb}  

\usepackage{float}
\usepackage{verbatim}
\usepackage{textcomp}
\usepackage{siunitx}
\usepackage{subcaption}
\usepackage{cite}
\usepackage{hyperref}
\usepackage{siunitx}
\usepackage{multirow}
\usepackage{color}
\usepackage{todonotes}
\renewcommand{\baselinestretch}{0.92}

\definecolor{mmcolored}{rgb}{0.0,0.0,0.0}
\newcommand{\mm}[1]{\textcolor{mmcolored}{#1}}

\title{\LARGE \bf
STPOTR: Simultaneous Human Trajectory and Pose Prediction Using a Non-Autoregressive Transformer for Robot Follow-Ahead
}

\author{Mohammad Mahdavian$^{*}$, Payam Nikdel$^{*}$, Mahdi TaherAhmadi, and Mo Chen
  \thanks{The authors are with School of Computing Science, Simon Fraser University (SFU), Burnaby, Canada {\tt\small \{mmahdavi, pnikdel, mtaherah, mochen\}@sfu.ca}}
  \thanks{This work received support from Amii and the CIFAR Program. M. Mahdavian received support from the SFU Graduate Deans Entrance Scholarship.}
  \thanks{* These authors contributed equally to this work}
}

\begin{document}

\maketitle
\thispagestyle{empty}
\pagestyle{empty}

\begin{abstract}

In this paper, we greatly expand the capability of robots to perform the follow-ahead task and variations of this task through development of a neural network model to predict future human motion from an observed human motion history.
We propose a non-autoregressive transformer architecture to leverage its parallel nature for easier training and fast, accurate predictions at test time. 
The proposed architecture divides human motion prediction into two parts: 
1) the~\textit{human trajectory}, which is the 3D positions of the hip joint over time, and 2) the~\textit{human pose} which is the 3D positions of all other joints over time with respect to a fixed hip joint.
We propose to make the two predictions simultaneously, as the shared representation can improve the model performance.
Therefore, the model consists of two sets of encoders and decoders. 
First, a multi-head attention module applied to encoder outputs improves human trajectory. 
Second, another multi-head self-attention module applied to encoder outputs concatenated with decoder outputs facilitates the learning of temporal dependencies.
Our model is well-suited for robotic applications in terms of test accuracy and speed, and compares favorably with respect to state-of-the-art methods. 
We demonstrate the real-world applicability of our work via the \textit{Robot Follow-Ahead} task, a challenging yet practical case study for our proposed model. 
The human motion predicted by our model enables the robot follow-ahead in scenarios that require taking detailed human motion into account such as sit-to-stand, stand-to-sit.
It also enables simple control policies to trivially generalize to many different variations of human following, such as follow-beside.
Our code and data are available at the following Github page: \url{https://github.com/mmahdavian/STPOTR}

\end{abstract}

\section{INTRODUCTION}
The recent advancements in robotics are making it possible to see an increase in the use of robots in daily life that require close human-robot interaction. This includes autonomous baggage, shopping carts, robots that help care for the elderly, and robots that can follow a human user while keeping close proximity. These applications are becoming closer to becoming a reality.  
In general, a robot can follow a human from behind~\cite{gockley2007natural,leigh2015person}, front~\cite{nikdel2021lbgp} and side by side~\cite{karunarathne2018model}. Following a user from behind or side by side is a well-studied subject. But there are only few studies that approach and give a solution for robot follow-ahead, as it is a much more difficult problem. However, several applications necessitate a robot follow-ahead approach due to security concerns. For instance, in the case of autonomous luggage, it can easily be stolen if it follows behind, and a study has shown that a robot following a human from behind can cause the user to frequently check that the robot is maintaining a safe distance and not hitting them~\cite{jung2012control}.
\begin{figure}[t]
  \centering
  \includegraphics[width=\columnwidth]{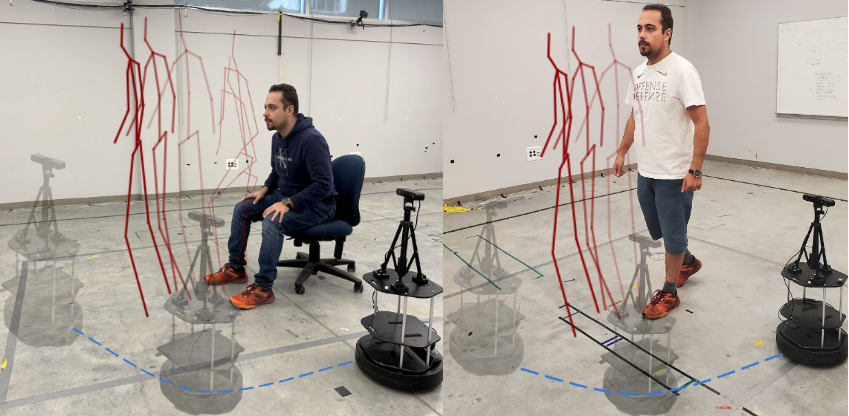} 
  \caption{Robot follow-ahead via human motion prediction}
  \label{cool1}
  \vspace{-6mm}

\end{figure}

To follow a target human from behind, one can simply use a proportional integral derivative (PID) controller to control the robot's heading to keep a human target in the middle of the image frame and use another one to keep a proper distance with respect to the target~\cite{jung2012control}. Also, for side-by-side following, Lidar sensors can help to keep a safe distance to a target~\cite{karunarathne2018model}. But following from the front side (follow-ahead) is a much more complicated process as it needs the robot to predict the human target's next intent.

In recent years, with the advancement of deep reinforcement learning (RL), a few works tried to solve the robot follow-ahead problem by implicitly learning the robot dynamics and predicting a person's future trajectory~\cite{nikdel2021lbgp}. Also, there are a few Kalman Filter-based methods that have tried to address this problem using classical control and estimation~\cite{ho2012behavior}. Other works have used image-based human trajectory prediction methods~\cite{huang2019stgat}. Despite the recent work, current robot follow-ahead capabilities leave much to be desired, as they are capable of predicting only human future trajectory while walking and crucially do not consider human body poses, which can convey an important information about human motion. 
For example, when a human decides to sit down or stand up or when a human is smoking or walking with a dog~\cite{nikdel2021lbgp}, human trajectory prediction alone may not be enough. In cases like this, human pose prediction is essential for acquiring a better understanding of the future motion and for solving the robot follow-ahead problem.

To account for human body pose in robot follow-ahead, accurate predictions of both future human 3D body pose and trajectory are needed. Given the predictions, the robot can move toward a point in front of the predicted future human pose. In the past, recurrent neural network (RNN) models have been utilized for predicting human 3D body pose. However, these models have been found to either not be accurate enough or too slow to be used in real-time robotic applications. They are also affected by "exposure bias"~\cite{schmidt2019Generalization}, which is caused by the accumulation of prediction errors over time due to the autoregressive nature of the models. These autoregressive models are more computationally intensive since the predicted elements are generated one at a time.
Transformers~\cite{vaswani2017attention} are a solution for these problems as they can be trained and tested~\cite{martinez2021pose} in parallel. Currently, very few works have attempted to jointly predict human pose and trajectory~\cite{nikdel2022}. Existing work tends to be either not accurate enough~\cite{yuan2020dlow} or very slow~\cite{nikdel2022}.

In this paper, we introduce an accurate and fast non-autoregressive transformer for simultaneous prediction of human trajectory and body poses. We demonstrate favorable results compared to previous works in terms of both speed and accuracy. Also, our ablation study shows the benefit of simultaneous human pose and trajectory prediction and a shared attention module in between that improves the model performance. During tests, we first estimate the human pose in each frame from captured images and use a sequence of the estimated frames to predict future ones. Finally, as shown in Fig.~\ref{cool1}, given the predicted human body motion, a robot trajectory planner moves the robot toward a point ahead of the human's predicted future state. In summary, our contributions are as follows:
\begin{itemize}
  \item Our approach to the robot follow-ahead task outperforms previous methods, and our results demonstrate the advantages of incorporating human body pose into this task. Specifically, our approach enables the robot to exhibit novel following behaviors that were not previously feasible, resulting in significant performance improvements.
  \item To the best of our knowledge, we are the first to simultaneously predict human pose and trajectory and utilize the results in a real-world robotic scenario.
  \item We achieve a reasonable accuracy for both human trajectory and body pose predictions with respect to the state-of-the-art methods.
  \item Using ablation studies, we show that our proposed shared attention module allows human body pose information to improve human trajectory prediction.
  \item We demonstrate our method in numerous human-following tasks on a real robot.
\end{itemize}

\section{Related Works}
\vspace{-1mm}
Human-following robots have been studied for ground~\cite{nikdel2018hands,pierre2018end}, aerial~\cite{huh2013integrated,jimenez2014framework} and underwater environments~\cite{nadj2020using}. For all cases, following from behind is the dominant scenario and arguably requires the least effort to develop. Classical human-following methods involve localizing the human and then at each step based on the human motion, find and navigate to the user's desired goal~\cite{jung2012control,wang2017real}. 

\subsection{Robot Follow-Ahead}

There are very few papers that have studied robot follow-ahead. In one of the first efforts, Ho et al.~\cite{ho2012behavior} assumed a nonholonomic human model and estimated human's linear and angular velocity via a Kalman filter. Their proposed motion planner did not perform well for some relatively complex scenarios. The authors in \cite{nikdel2018hands} developed an Extended Kalman Filter (EKF) approach by combining 2D Lidar and a fish-eye camera to detect and track a person. A velocity-based heading estimator and human model that accounted for obstacles helped to correct the EKF predicted position. More recently, Nikdel et al. used Deep RL and curriculum learning to learn a robust policy for robot follow-ahead \cite{nikdel2021lbgp}. However, the latter two methods did not account for human body pose and thus are limited in their use case. For example, the EKF method cannot perform well when the human is nearly stationary, and the deep RL method requires ground truth human position and heading via motion capture.

\begin{figure*}[h]
\vspace{3mm}
  \centering
  \includegraphics[width=0.88\textwidth]{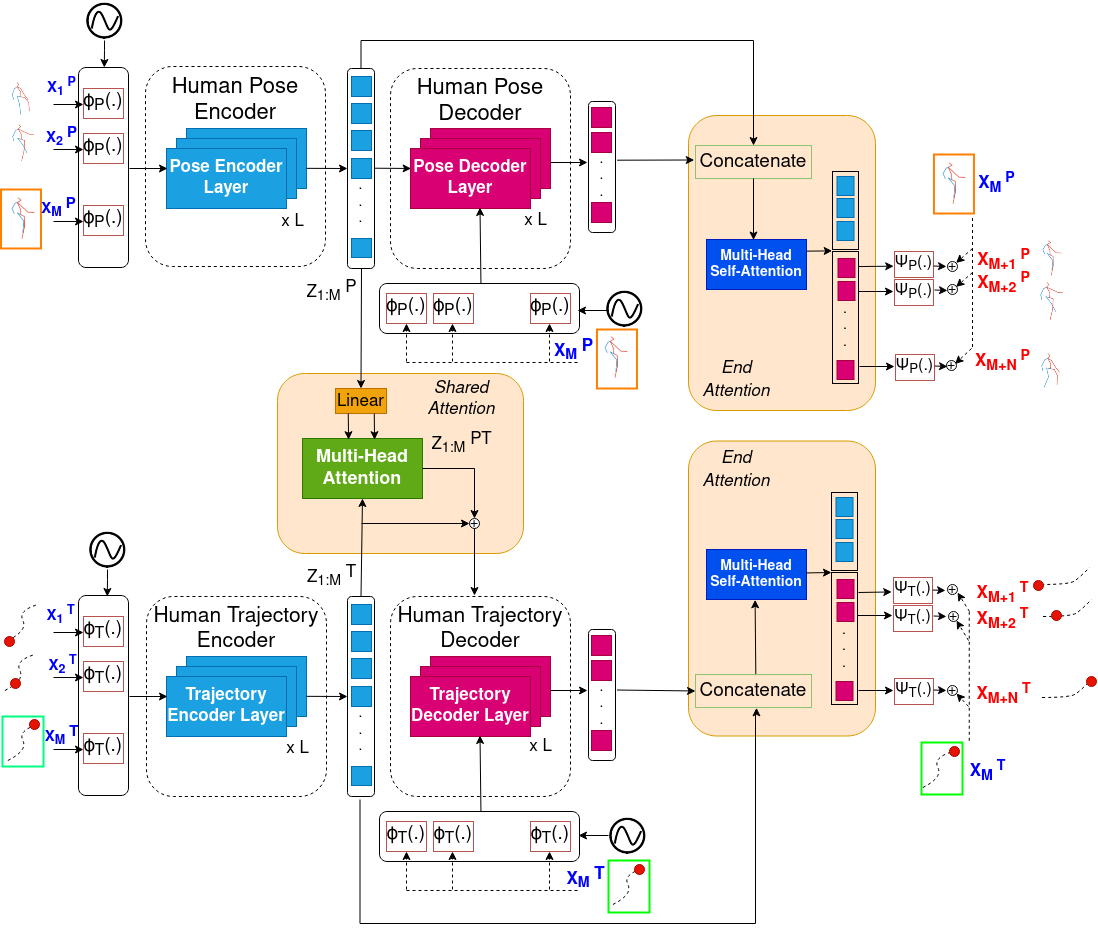} 
  \caption{Our model architecture predicts both human poses and trajectories concurrently based on an observed 3D human joint sequence. It consists of two non-autoregressive transformers for pose and trajectory predictions, with a shared Attention module that enhances the quality of predictions by facilitating the exchange of knowledge between the two. To model the temporal dependencies more effectively, an End Attention module is added to the end of each decoder. The blue-colored frames show the input sequence or frame and the red ones show the output. The rectangular frames show that the same frame (last input pose) is copied and used as the decoder input sequence and as a residual for decoder output.}
  \label{my_model}
  \vspace{-5mm}
  \end{figure*}

\subsection{Human Pose Prediction}

In general, human pose prediction methods can be categorized into probabilistic or deterministic~\cite{lyu20223d}. In probabilistic methods, similar to how our brains function, the goal is to predict multiple scenarios from an observed sequence of frames. 
Currently, the state-of-the-art work in this area is Diversifying Latent Flows (DLow), which predicts multiple hypotheses for human 3D poses using pre-trained deep generative models. We use this method as one of our baselines for human pose prediction. Another recent work that takes a very different probabilistic approach -- just accounting for human trajectory and not pose -- predicts the probability distribution over human kinematic states~\cite{agand2022human}.
In deterministic approaches, one predicts a human pose sequence~\cite{aksan2021spatio} or trajectory sequence~\cite{zhang2021multimodal} from the observed motion. 

Many deep learning approaches use RNN-based models to predict human motion~\cite{fragkiadaki2015recurrent}; however, these autoregressive models have two major shortcomings. First, they are prone to accumulative prediction errors and second, they are not parallelizable which makes them computationally intensive for testing~\cite{martinez2021pose}. Recently, a few methods tried to prevent the drift issue by including adversarial losses and enhance prediction quality with geodesic body measurements~\cite{gui2018adversarial} which makes them more difficult to train and stabilize. Also, to better embed the joint dependencies, some methods combine their algorithm with spatio-temporal modeling to better learn the relation between all the joints in a single frame or a sequence of frames~\cite{li2021multiscale,fu2022learning}.

With the improvement of transformer models~\cite{vaswani2017attention}, in a few studies, they have been employed  to solve the human pose prediction problem. Aksan et al. proposed an autoregressive transformer to learn to decouple spatio-temporal representations~\cite{aksan2021spatio}. They achieve acceptable results in term of accuracy; however, the autoregressive nature caused the algorithm to be slow. On the other hand, González et al. developed a non-autoregressive version of the transformers called Pose Transformer (POTR), which can perform faster with lower accuracy~\cite{martinez2021pose}. They use the main encoder-decoder structure of transformers~\cite{vaswani2017attention} to learn the temporal dependencies. They also use Graph Convolutions Network (GCN) and Multi-Layer Perceptron (MLP) based layers in their encoder and decoder networks, respectively, to learn the spatial dependencies between all joints in one frame. During training and testing, the last observed frame is copied and used as their decoder input and with a residual connection to the decoder output. Therefore the decoder would learn the sequence offset with respect to the last seen frame. The design of our model is partially inspired by Pose Transformer (POTR)~\cite{martinez2021pose}. However, all mentioned methods consider a fixed hip joint and even in some cases fixed heading, which makes them impractical for many robotic applications. Therefore, we have made multiple improvements to the model and data structure to make it suitable for the human-following task.
\vspace{-2mm}
\subsection{Human Trajectory Prediction}

Many human trajectory predictions have been developed for autonomous driving systems~\cite{postnikov2021transformer,giuliari2021transformer,achaji2022pretr}, in which the primary goal is to predict the future trajectory of pedestrians in order to avoid colliding with them. Recently, some works attempted to use transformers~\cite{vaswani2017attention} to predict multiple possible human trajectories, but very few methods attempted to simultaneously predict human poses and trajectory. DMMGAN~\cite{nikdel2022} performs this task with reasonable accuracy, at the cost of lower speed. We use this paper as another baseline for our method, and show that our method is preferred for robotic purposes due to faster inference speed. 
\vspace{-1mm}
\section{Problem Formulation}
Given a length-$M$ sequence of global human 3D joint positions (skeleton) $X_{1:M}$, we seek to predict a length-$N$ sequence of future human 3D joint positions $X_{M+1:M+N}$. Each $X_i\in\mathbb{R}^{51}$ represents one frame of seventeen 3D global human joint positions at frame $i$. Fig.~\ref{cool1} illustrates the problem of 3D global joint position prediction.

\section{Methodology}
\vspace{-1mm}
We divide the prediction task into two interdependent parts. 
The first part is to predict the future 3D hip trajectory, $X_{M+1:M+N}^T$, from previously observed ones, $X_{1:M}^T$. 
The hip is the standard joint position for representing the 3D human position purpose~\cite{postnikov2021transformer,giuliari2021transformer,achaji2022pretr}.
Next, as the second part of the problem, we aim to predict a future 3D human pose sequence, $X_{M+1:M+N}^P$, from the observed ones, $X_{1:M}^P$. Here a 3D human pose is defined as all joints' relative 3D position with respect to the fixed hip joint. 
The superscript $T$ and $P$ denote the human trajectory and pose sequence, respectively. 
We aim to solve the two parts simultaneously, as the features transferred in between can improve the predictions. 

In this paper, we propose to solve this problem by conditional sequence modeling where the goal is to train the set of parameters of a non-autoregressive transformer. 

\subsection{Human Body Motion Prediction}

In our model, we follow the main structure of the autoregressive~\cite{vaswani2017attention} and non-autoregressive~\cite{martinez2021pose} transformers with multiple improvements and adjustments. 
Fig.~\ref{my_model} shows the structure of our model architecture. 
The model simultaneously predicts the human pose (upper section) and trajectory (lower section). 
The encoders and decoders are composed of \textit{L} layers, each with the structure in~\cite{vaswani2017attention}, containing multi-head, self- or encoder-decoder attention layers as well as fully-connected layers.
The encoders receive a sequence of 3D human poses $X_{1:M}^{P}$ or hip trajectory $X_{1:M}^{T}$, and generate the two sequences of embeddings $Z_{1:M}^P$ and $Z_{1:M}^T$. 
While the main structure of the transformer model learns the temporal dependencies, two networks are added ($\phi$ and $\psi$) as pose encoder (GCN-based) and pose decoder (MLP-based) to identify the spatial dependencies between the joints in each frame. 
The pose and trajectory encoding networks, $\phi_P$ and $\phi_T$, are GCNs that learn the spatial relationship between the body joints. 
The weight of the graph edges represented by the adjacency matrix is used to compute embeddings of dimension \textit{D} for the human pose and human trajectory vectors in the input sequences $X_{1:M}^{P}$ and $X_{1:M}^{T}$.
In order to modify the model to perform in a non-autoregressive manner, the last frame of input sequences, $X_M^{P}$ and $X_M^{T}$, were copied and used as \textit{query sequences} for decoders input. 
The model generates pose and trajectory predictions $X_{M+1:M+N}^{P}$ and $X_{M+1:M+N}^{T}$, in parallel using the networks $\psi_P$ and $\psi_T$, from the decoder outputs and a residual connection containing the query sequences. 
Therefore, the decoders learn the offsets with respect to the last seen frame.

One of the benefits of our architecture is that we can share the representation between human pose and trajectory prediction modules. 
In order to fully benefit from the combination of human poses and hip trajectory, we have added a multi-head attention module called~\textit{Shared Attention} to apply attention between pose and trajectory encoder outputs as shown in the middle of Fig. \ref{my_model}.
First, we apply a linear layer to the pose encoder embedding, $Z_{1:M}^P$, to change the dimension from pose to trajectory embedding size. 
Then, we pass it with a copy as well as the trajectory encoder embedding, $Z_{1:M}^{T}$, to the multi-head attention module. 
We then add the multi-head attention output, $Z_{1:M}^{PT}$, with the hip trajectory encoder output to use it in the hip trajectory decoder.
The added multi-head attention can improve the hip trajectory prediction compared to solely relying on hip trajectory history, since the human pose changes are related to how humans move overall. 
In Section~\ref{subsec:ablation} we investigate how this attention module can help our model predict more accurately.

In addition, we have added a multi-head attention layer to the end of each decoder called~\textit{End Attention}. 
This module can help the model to better learn the temporal dependencies between all frames. 
We concatenate the pose and trajectory encoders' output with the decoders' output and apply a self-attention module. 
Then we output the last encoded features with the same length as the target sequence length. 
To convert them to the actual sequence of future 3D human pose $X_{M+1:M+N}^P$ and hip trajectory $X_{M+1:M+N}^T$, the model uses a pose and trajectory decoder ($\psi$). 
We discuss the impact of this module in the ablation study presented in Section~\ref{subsec:ablation}.


\section{Human Motion Prediction Experiments}
In this section, we first describe the dataset used to train our model, implementation details, baselines, and metrics. 
Then, we show the performance of our human motion prediction method with respect to baselines.
Finally, we present the results of ablation studies to demonstrate the effectiveness of different parts of our proposed architecture.

\subsection{Preliminaries and Implementation Details}
\subsubsection{Dataset}
To train the human motion prediction model, we used the well-known and standard Human3.6M dataset~\cite{ionescu2013human3}. 
It contains the 3D joint position of seven actors performing 15 activities, including walking, sitting, and smoking. 
Traditionally, this dataset has been used as a benchmark for human pose prediction~\cite{lyu20223d}, but we utilize it for human trajectory prediction as well. 
As explained before, we extracted the hip trajectory of each actor for the human trajectory prediction and all other joints' relative positions with respect to the fixed hip for human pose prediction. 
Conventionally, for this dataset, one reduces the frame rate from 50 Hz to 25 Hz~\cite{yuan2020dlow,aksan2021spatio,martinez2021pose}; however, we used 10 Hz, a more suitable frame rate for robotic purposes as it is fast enough, reduces the complexity of our model, and speeds up predictions at test time. 
Also, we followed the standard input and output duration of our human pose prediction baseline, DLow~\cite{yuan2020dlow} which are 0.5 sec (5 frames) for input and 2 sec (20 frames) for the output.

\begin{table}[t]
\vspace{2mm}
\begin{center}
\caption{Analytical comparisons between our developed model and the baselines introduced in~\cite{yuan2020dlow} and~\cite{nikdel2022} in terms of~\textit{ADE} and~\textit{FDE} for both human pose and trajectory predictions and Inference Duration (ID)}
\label{table_1}
\begin{tabular}{c@{\hspace{1.5\tabcolsep}}c@{\hspace{1.5\tabcolsep}}c@{\hspace{1.5\tabcolsep}}c@{\hspace{1.5\tabcolsep}}c@{\hspace{1.5\tabcolsep}}c}
Method & $ADE_{Pose}$ & $FDE_{Pose}$ & $ADE_{Traj}$ & $FDE_{Traj}$ & $ID$\\
       & (m)  & (m)  & (m)  & (m) & (msec)\\
\hline
DLow~\cite{yuan2020dlow} & 0.48 & 0.62 & 0.19 & 0.45 & 20\\
DMMGAN~\cite{nikdel2022} & 0.44 & 0.52 & 0.12 & 0.23 & 100\\
HipOnly~\cite{nikdel2022} & NA & NA & 0.15 & 0.30 & 18\\
Ours & 0.50 & 0.75 & 0.13 & 0.27 & 25\\
\end{tabular}
\end{center}
\vspace{-8mm}
\end{table}

\subsubsection{Training}

We used Pytorch as our deep-learning framework. 
The model was trained with AdamW~\cite{loshchilov2019decoupled} for 250 epochs with a learning rate of $10^{-4}$ and a batch size of 16. 
The model was trained after 50K steps with warm-up 
scheduled in the first 10K steps. During warm-up, the learning rate gradually increases from zero to $10^{-4}$, which increases training stability.

\subsubsection{Model Hyperparameters
}

Based on experience, we set the embedding dimensions to $D_{Pose}=512$ for pose prediction and $D_{Traj}=64$ for trajectory prediction. 
Also, the fully-connected dimension in our encoders and decoders was set to 2048. 
The encoders and decoders each contain four layers of pre-normalized~\cite{xiong2020layer} multi-head attention modules with eight attention heads. 
Here, ``pre-'' or ``post-normalized'' refers to whether the normalization layer is the first layer in the multi-head attention module or the last one.

\subsubsection{Baselines}

As our baselines, we compared our work with two state-of-the-art works in human pose and trajectory predictions suitable for robotic purposes. We used DLow~\cite{yuan2020dlow} as our first baseline as a fast and accurate method in human pose prediction. This method has the best performance for pose prediction out of all other methods except for DMMGAN~\cite{nikdel2022}. Since this method only predicts human poses at 25 Hz, we retrained it for simultaneous human pose and hip trajectory predictions at 10 Hz with hip joint motion added to the predictions to be able to compare directly. 
As a more accurate but slower method, we compared our results with DMMGAN~\cite{nikdel2022} that simultaneously predicts human pose and trajectory for robotic purposes. 
As another baseline for trajectory predictions, we compare our method with a simple GRU-based method called~\textit{Hip Only} introduced as a trajectory prediction baseline in~\cite{nikdel2022}. 
In this baseline, a GRU is applied to the human trajectory after passing through a transformer encoder.
To the best of our knowledge, these are the only available methods to compare with simultaneous human pose and trajectory predictions suitable for real-world robotic purposes. Other prior methods -- and DLow~\cite{yuan2020dlow} without any modifications -- either only predict human body pose relative to the fixed hip or heading~\cite{aksan2021spatio,martinez2021pose} without predicting the hip trajectory in 3D space or are not fast enough.
\subsubsection{Metrics}

In order to compare our results with the baselines, we use the conventional Average Displacement Error (ADE) and Final Displacement Error (FDE)~\cite{lyu20223d} metrics. 
ADE is the average of the $L_2$ distance over all time steps between ground truth and prediction. 
FDE is the $L_2$ distance between the last ground truth and predicted frames. 
We compared both metrics for both pose and trajectory predictions. 
As another important factor for real-time robotic purposes, we compared the algorithms' speed at test time. 

\subsection{Main Results}

Table~\ref{table_1} quantitatively compares our method to the baselines. 
The achieved $ADE_{Pose}$ is comparable to the state-of-the-art DLow~\cite{yuan2020dlow} paper and DMMGAN~\cite{nikdel2022}. 
Also, we have achieved better results for trajectory prediction with respect to DLow and~\textit{Hip Only}~\cite{nikdel2022}. All training and testing were done on a laptop with an Intel CPU Core i9-9980HK CPU and RTX 2080 Max-Q GPU. Due to the non-autoregressive nature of our method, we were able to achieve a much better computation speed at test time compared to DMMGAN, and similar computation speed compared to DLow.
However, our method has slightly worse but comparable $ADE_{Pose}$ and $FDE_{Pose}$ with respect to DLow and $ADE_{Traj}$ and $FDE_{Traj}$ with respect to DMMGAN~\cite{nikdel2022}. 
This result was expected as discussed in~\cite{martinez2021pose}: The non-autoregressive nature of the model reduces the model's capability in modeling correlation between frames which increases model error. Another reason is that DLow and DMMGAN predict multiple possible predictions for an input sequence and $ADE_{Pose}$ and $FDE_{Pose}$ are calculated for the most similar predicted sequence to the ground truth; thus, they are somewhat similar to ensemble methods in spirit.

\begin{figure*}[h]
\vspace{3mm}
  \centering
  \includegraphics[width=0.85\textwidth]{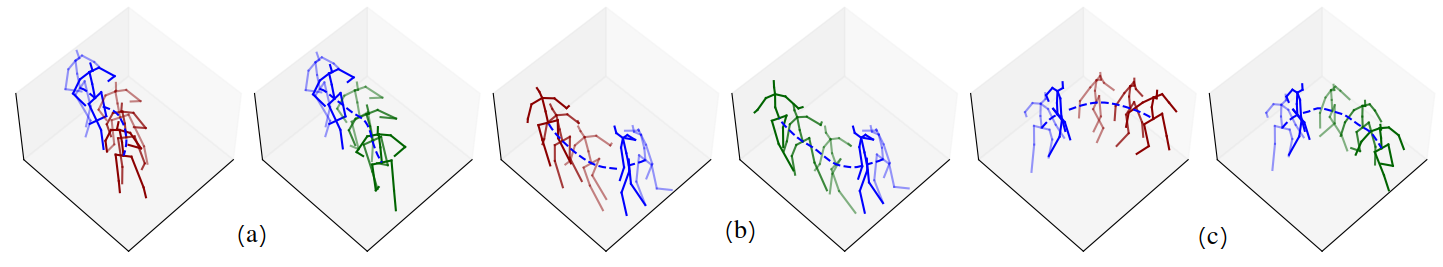} 
  \caption{Three samples of the predicted motion vs. ground truth. On each couple of figures (a to c) the left one shows the predicted motion given an observed sequence and the right one shows the ground truth. The blue-colored skeletons show the input sequence and the red and green ones show the model predictions and ground truth, respectively. Also, the trajectory of the hip is shown with dashed \mm{blue} lines.}
  \label{motion2}
  \vspace{-3mm}

\end{figure*}

Note that for our robotic follow-ahead task, to work smoothly, we need to make the predictions, pre-processing (3D human pose estimation) and post-processing (robot trajectory planning) in less than 100 msec as the frame rate of the model input is 10 Hz. Therefore, the DMMGAN~\cite{nikdel2022} was not a suitable choice for this task. On the other hand, DLow's trajectory prediction accuracy was not adequate. Therefore, our method provided the most suitable model in terms of both accuracy and speed. For robotic purposes, our accuracy is adequate as demonstrated in section~\ref{follow_ahead_results}, and the fast computation speed at test time as needed. Also, Fig.~\ref{motion2} qualitatively shows three samples of the predicted motion with respect to the ground truth.





\begin{table}[b]
\vspace{-2mm}
\begin{center}
\caption{Our ablation study analytical comparisons}
\label{table_2}
\begin{tabular}{ c@{\hspace{1.5\tabcolsep}}  c@{\hspace{1.5\tabcolsep}}  c @{\hspace{1.5\tabcolsep}} c @{\hspace{1.5\tabcolsep}} c }
Model & $ADE_{Pose}$ & $FDE_{Pose}$ & $ADE_{Traj}$ & $FDE_{Traj}$\\
& (m) & (m) & (m) & (m)\\
\hline
\textit{No Shared Attn}  & 0.50 & 0.75 & 0.16 & 0.33 \\
\textit{Shared Attn-Pose Only}  & 0.51 & 0.76 & 0.16 & 0.33 \\
\textit{No End Attn}  & 0.52 & 0.77 & 0.18 & 0.33 \\
\textit{Post Normalized} & 0.51 & 0.76 & 0.17 & 0.32 \\
Ours & \textbf{0.50} & \textbf{0.75} & \textbf{0.13} & \textbf{0.27} \\
\end{tabular}
\end{center}
\vspace{-1mm}

\end{table}

\subsection{Ablation Study}
\label{subsec:ablation}
We performed an ablation study to evaluate the training process and the effectiveness of different modules used in our model. 
To show one of the major advantages of our method, we discuss the effect of the~\textit{Shared Attention} module used for better trajectory predictions. 
We compare the current results with the cases that 1) no such module is applied (\textit{No Shared Attn}) and 2) the shared attention module is applied only for pose predictions (\textit{Shared Attn-Pose Only}).
Also, we study the effect of the~\textit{End Attention} module added to the end of each decoder which aims to better model temporal dependencies by removing this module (\textit{No End Attn}).
Finally, we compare the achieved accuracy with the post-normalized~\cite{xiong2020layer} multi-head attention modules. 

Based on Table~\ref{table_2}, the shared attention module has improved the trajectory prediction by incorporating the human pose representation while predicting trajectory. The same module degraded the pose prediction \mm{(\textit{Shared Attn-Pose Only})} and we believe there are two reasons for it. First, in some of the dataset motions, the body limbs have random movements, such as random hand waving during walking, which makes the predictions harder. Second, while the body pose can be informative for predicting the hip trajectory, the reverse may not be true, as given a hip trajectory, there are often still a lot of degrees of freedom for the body pose. Also, the end attention module applied to the concatenation of encoder and decoder outputs improved the model performance by better modeling the temporal dependencies between input and output frames. In addition, the post-normalized structure for multi-head attentions was not able to perform as well as the current pre-normalized version.

\begin{table}[h]
  \centering
  \caption{Robot follow-ahead comparative results for three tested scenarios.}
  \label{table_3}
  \begin{tabular}{ c @{\hspace{1\tabcolsep}} |l| r |c|l|r}
    Human & Method & Reward &  Human & Method & Reward  \\
     Trajectory& && Trajectory&  \\\hline
    \multirow{5}{*}{\begin{minipage}{.12\linewidth}
      \includegraphics[width=0.95\linewidth]{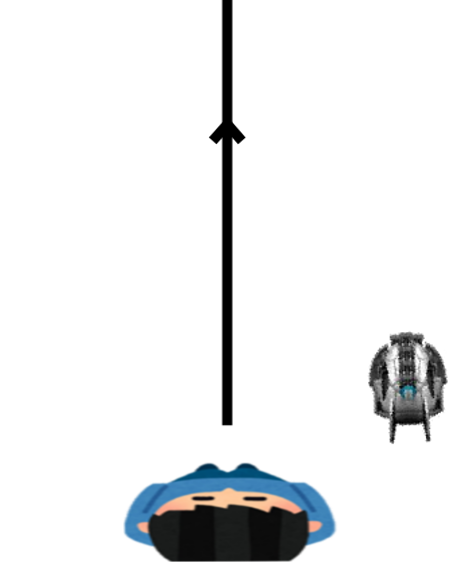}
    \end{minipage}}
    & & &    \multirow{5}{*}{\begin{minipage}{.12\linewidth}
      \includegraphics[width=0.95\linewidth]{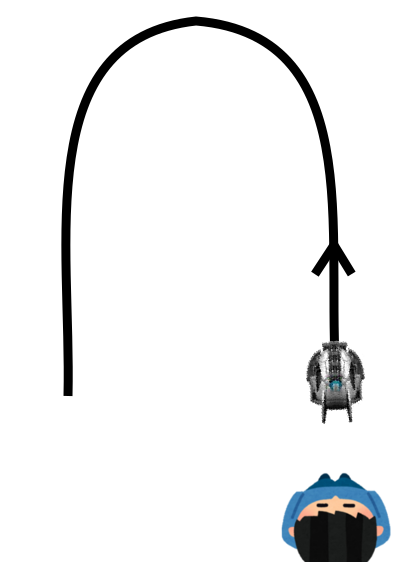}
    \end{minipage}} && \\ 
    & Ours & 25.94 & & Ours & 13.11\\
    &LBGP & \bm{$27.42$} &    &LBGP & \textbf{14.91}\\
    &HC & $26.31$ &&HC & $-11.33$\\
    &&&&&\\
    \hline
    \multirow{5}{*}{\begin{minipage}{.12\linewidth}
      \includegraphics[width=0.8\linewidth]{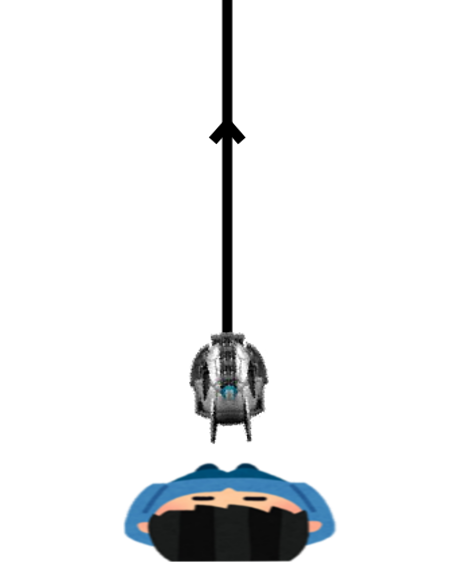}
    \end{minipage}}
    & & &\multirow{5}{*}{\begin{minipage}{.12\linewidth}
      \includegraphics[width=0.8\linewidth]{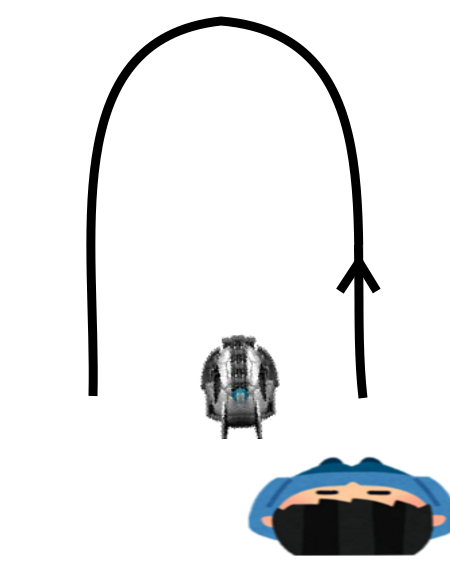}
    \end{minipage}} & &\\
    & Ours & 25.31 && Ours & 8.89 \\ 
    &LBGP & $40.92$& &LBGP & \textbf{18.73}\\
    &HC & \bm{$61.08$} & &HC & $-8.09$ \\
     &&&&&\\
    \hline
    \multirow{5}{*}{\begin{minipage}{.12\linewidth}
      \includegraphics[width=0.95\linewidth]{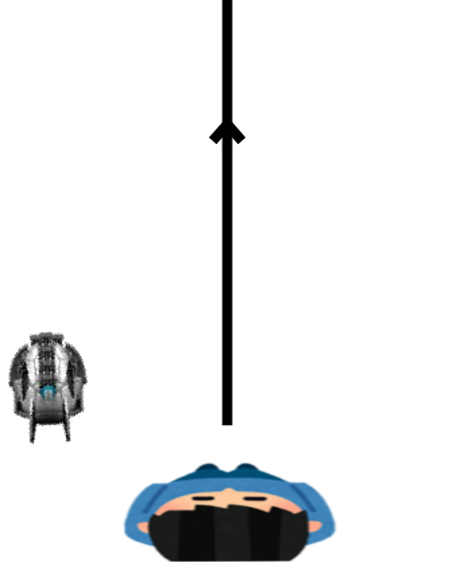}
    \end{minipage}}
    & &     &\multirow{5}{*}{\begin{minipage}{.12\linewidth}
      \includegraphics[width=0.95\linewidth]{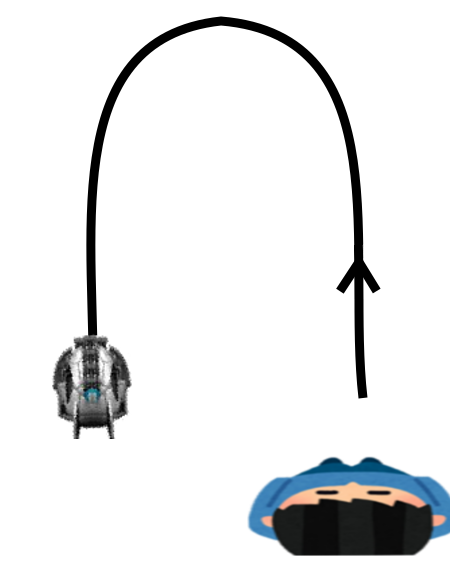}
    \end{minipage}} & &\\
    & Ours & 7.62&    & Ours & $-10.92$ \\
    &LBGP & \bm{$15.68$} & &LBGP & \bm{$16.52$} \\
    &HC & $14.24$&    &HC & $-13.14$ \\
        &&&&&\\
    \hline
    \multirow{5}{*}{\begin{minipage}{.12\linewidth}
      \includegraphics[width=0.95\linewidth]{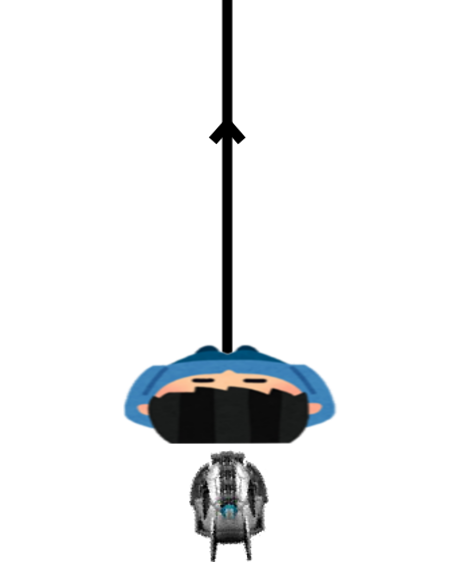}
    \end{minipage}}
    & &&\multirow{5}{*}{\begin{minipage}{.12\linewidth}
      \includegraphics[width=0.95\linewidth]{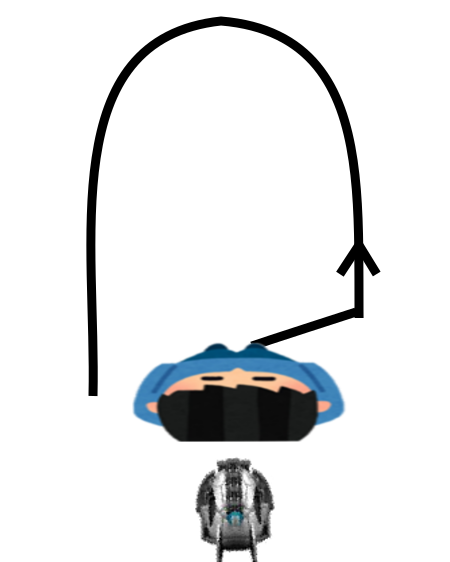}
    \end{minipage}}
    & &\\ 
    &Ours & \bm{$-5.14$} &&Ours & $7.37$  \\
    &LBGP & $-6.79$ &&LBGP & \bm{$20.03$}  \\
    &HC & $-15.69$ &    &HC & $-11.83$\\
    &&&&&\\
    
    \hline
    \multirow{5}{*}{\begin{minipage}{.12\linewidth}
      \includegraphics[width=0.95\linewidth]{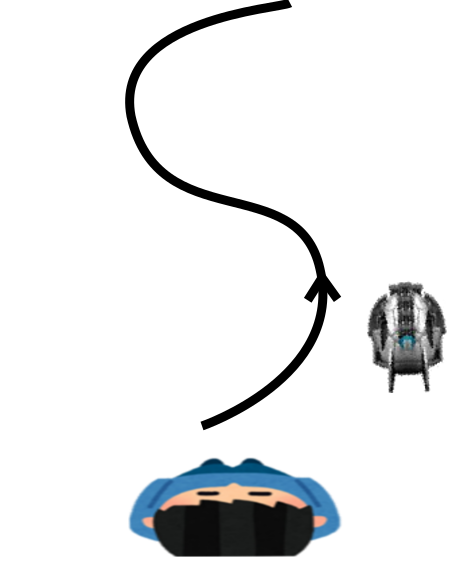}
    \end{minipage}}
    & &     &\multirow{5}{*}{\begin{minipage}{.12\linewidth}
      \includegraphics[width=0.95\linewidth]{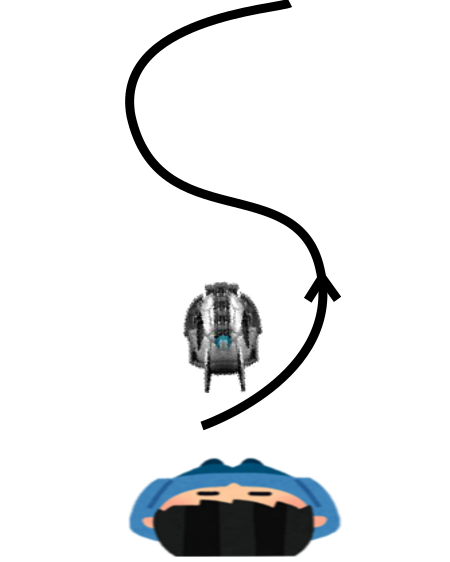}
    \end{minipage}} & &\\
    & Ours & \textbf{22.02}&    & Ours & \textbf{24.15} \\
    &LBGP & 2.30 & &LBGP & 5.15 \\
    &HC & $-9.42$&    &HC & $-6.83$ \\
        &&&&&\\
    \hline
    \multirow{5}{*}{\begin{minipage}{.12\linewidth}
      \includegraphics[width=0.95\linewidth]{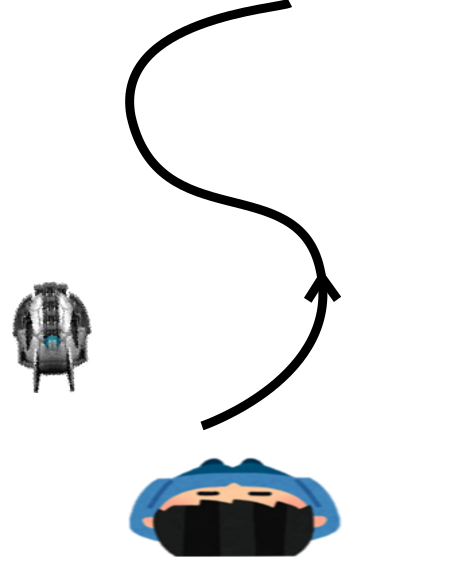}
    \end{minipage}}
    & &&\multirow{5}{*}{\begin{minipage}{.12\linewidth}
      \includegraphics[width=0.95\linewidth]{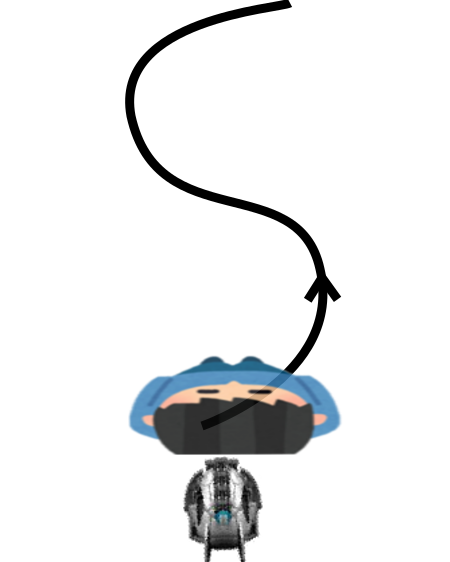}
    \end{minipage}}
    & &\\ 
    &Ours & \textbf{14.33} &&Ours & \textbf{8.28}  \\
    &LBGP & 11.74 &&LBGP & 7.84  \\
    &HC & $-9.29$ &    &HC & $-17.64$\\
\vspace{-4mm}

  \end{tabular}

\end{table}


\section{Robot Follow-Ahead}
In this section, we demonstrate the real-world application of our proposed architecture. 
We use the trained model to perform the challenging task of robot follow-ahead and compare our method's performance with a hand-crafted (HC)~\cite{nikdel2018hands} and RL-based (LBGP)~\cite{nikdel2021lbgp} methods. We test the same scenarios in the two baselines that are follow-ahead on a~\textit{Straight line} and~\textit{S-Shaped} and~\textit{U-Shaped} curves. Similarly, the robot starting points are on four sides of the user's initial location (left, right, in front, and behind). In order to evaluate the follow-ahead task, a reward function similar to~\cite{nikdel2021lbgp} based on the relative position of the robot and the human is used. 

\subsection{Robot Follow-Ahead via Human Motion Predictions}
To use the trained model for the robot follow-ahead task in the real world, it was retrained with Gaussian noise added to the input sequence to improve the robustness to noisy inputs. A ZED2 camera was used as a 3rd person viewer for human pose estimation, and a turtlebot2~\cite{singh2018turtlebot} robot was used as the testing platform. We used a 3rd person camera to abstract away hardware complications such as limited field of view. Our focus is on demonstrating that the trained model for human motion prediction is useful for the follow-ahead task, and will leave hardware implementation details out of the scope of this paper. 
At each moment, the ZED2 camera captured an image and estimated the current human body 3D joints' positions. Then we sent the last 0.5 second frames (5 frames) to the STPOTR model and predicted the user's motion in the next 2 seconds (20 frames). We calculated the future human heading using the line created by the left and right hip joints positions on the 20th frame, and chose the point 1.5 meters in front, oriented in the same direction, as the robot goal pose (position and orientation). Then we used the Time Elastic Band (TEB) trajectory planner \cite{wu2021improved} to move the robot toward the goal pose. This planner also helps our follow-ahead system to stay a safe distance from the person at all times.

\subsection{Real-World Experimental Results}\label{follow_ahead_results}
\begin{figure}[t]
\vspace{-2mm}

\centering

\subfloat{
	\label{fig:str}
	\includegraphics[width=0.15\textwidth]{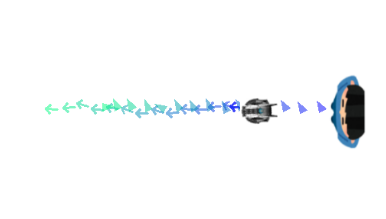} } 
\subfloat{
	\label{fig:S}
	\includegraphics[width=0.15\textwidth]{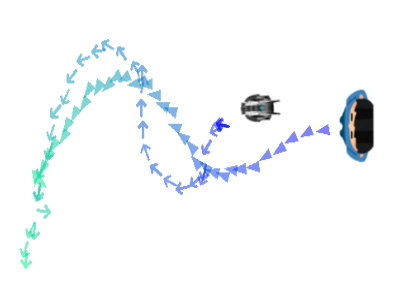}} 
\subfloat{
	\label{fig:U}
	\includegraphics[width=0.15\textwidth]{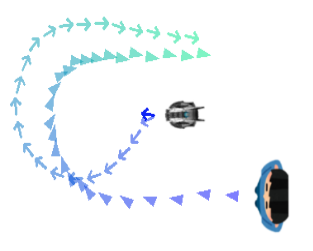}} 
\caption{Three samples of the robot follow-ahead tasks for U-Shaped, S-Shaped and straight line scenarios. The triangle and arrows show the human and robot motions, respectively. }
\label{follow-ahead-samples}
\vspace{-5mm}
\end{figure}

Table~\ref{table_3} compares the achieved reward by our robot follow-ahead method with respect to our baselines~\cite{nikdel2018hands,nikdel2021lbgp}. The reported rewards are the average reward values of two tests on two different users. 
We achieve a much higher reward in the~\textit{S-Shaped} scenarios which is a complicated motion and comparable results in other ones. Our method only performed poorly when the robot was placed on the human's left side during~\textit{U-Shaped} motion which can be due to the far distance between the human and robot during the initial periods of the motion. Fig.~\ref{follow-ahead-samples} shows a few of the follow-ahead motions in different scenarios. 

Crucially, note that the LBGP~\cite{nikdel2021lbgp} used a motion capture system for localizing the human and robot, which greatly simplifies the human-following problem, whereas we present a more realistic method that uses the much noisier 3D human pose estimation of the ZED2 camera. 

Perhaps most importantly, we were also able to account for much more detail in the human motion for follow-ahead task \mm{by simultaneously predicting the human's future pose and trajectory.} This enables our system to easily generalize to many different scenarios involving different human motions such as \textit{Sit-to-Stand} and \textit{Stand-to-Sit}, as well as to different variations of human following such as \textit{follow-beside} or keeping a \textit{variable distance} with the human depending on the human walking speed or surrounding environment. These are very difficult if not impossible scenarios and tasks for our baselines. For example, the RL-based LBGP~\cite{nikdel2021lbgp} method would require retraining of the policy for every variation of the human-following task. LBGP also does not account for the human body pose, and the HC~\cite{nikdel2018hands} method in addition does not consider the human heading. The application of our algorithm in these different scenarios and task variations can be found at \url{https://www.youtube.com/playlist?list=PLuLzEWWNu1_p1bjUHhWUHRMFOLmLpUCpM}

\vspace{-2mm}

\section{CONCLUSIONS}
In this paper, we presented simultaneous human trajectory and motion prediction for a real-world robotic purpose. We used two parallel non-autoregressive transformers and modified them for our purpose. We achieved a reasonable performance in terms of speed, pose, and trajectory prediction with respect to all the baselines which makes our method suitable for robotic purposes. We demonstrated our model capability by testing on the robot follow-ahead task and achieved better or comparable results with respect to previous methods. In future work, we will attempt to solve the robot follow-ahead problem using multiple onboard cameras to observe the human from a closer range in all views. 






\bibliographystyle{IEEEtran}
\bibliography{main}

\end{document}